\documentclass{article}
\usepackage[preprint,nonatbib]{neurips_2025}

\usepackage[utf8]{inputenc}
\usepackage[T1]{fontenc}

\usepackage[dvipsnames]{xcolor}
\definecolor{linkColor}{RGB}{237,4,140}
\definecolor{citecolor}{RGB}{0, 113, 188}
\usepackage[colorlinks=true,linkcolor=linkColor,citecolor=citecolor,filecolor=linkColor,urlcolor=linkColor]{hyperref}       

\usepackage{url}            
\usepackage{booktabs}       
\usepackage{amsfonts}       
\usepackage{nicefrac}       
\usepackage{microtype}      
\usepackage{makecell}
\usepackage{bbding}
\usepackage{wrapfig}

\usepackage{graphicx}
\usepackage{amsmath}
\usepackage{amssymb}
\usepackage{multicol}
\usepackage{multirow}
\usepackage{amsthm}
\usepackage{mathrsfs}
\usepackage{pifont}
\usepackage{diagbox}
\usepackage[ruled]{algorithm2e}
\usepackage{bm}
\usepackage{subfigure}
\usepackage{enumerate}
\usepackage{makecell}
\usepackage{enumitem}

\newcommand*{\system}{SimpleAR\@\xspace}
\newcommand*{\eg}{\emph{e.g.}\@\xspace}
\newcommand*{\etc}{\emph{etc}\@\xspace}
\newcommand*{\ie}{\emph{i.e.}\@\xspace}

\makeatletter
\newcommand\figcaption{\def\@captype{figure}\caption}
\newcommand\tabcaption{\def\@captype{table}\caption}
\makeatother

\title{\system: Pushing the Frontier of \\ Autoregressive Visual Generation through \\ Pretraining, SFT, and RL}
\author{Junke Wang$^{1}$,~Zhi Tian$^{2}$,~Xun Wang$^{2}$,~Xinyu Zhang$^{2}$,~Weilin Huang$^{2}$,~Zuxuan Wu$^{1}$,~Yu-Gang Jiang$^{1}$ \\
\\
$^{1}$Fudan University, $^{2}$ByteDance Seed \\
\\
Code is available at \href{https://github.com/wdrink/SimpleAR}{https://github.com/wdrink/SimpleAR}.
\vspace{-0.3in}
}

\begin{document}
\maketitle

\begin{figure*}[!ht]
\centering
\includegraphics[width=0.9\linewidth]{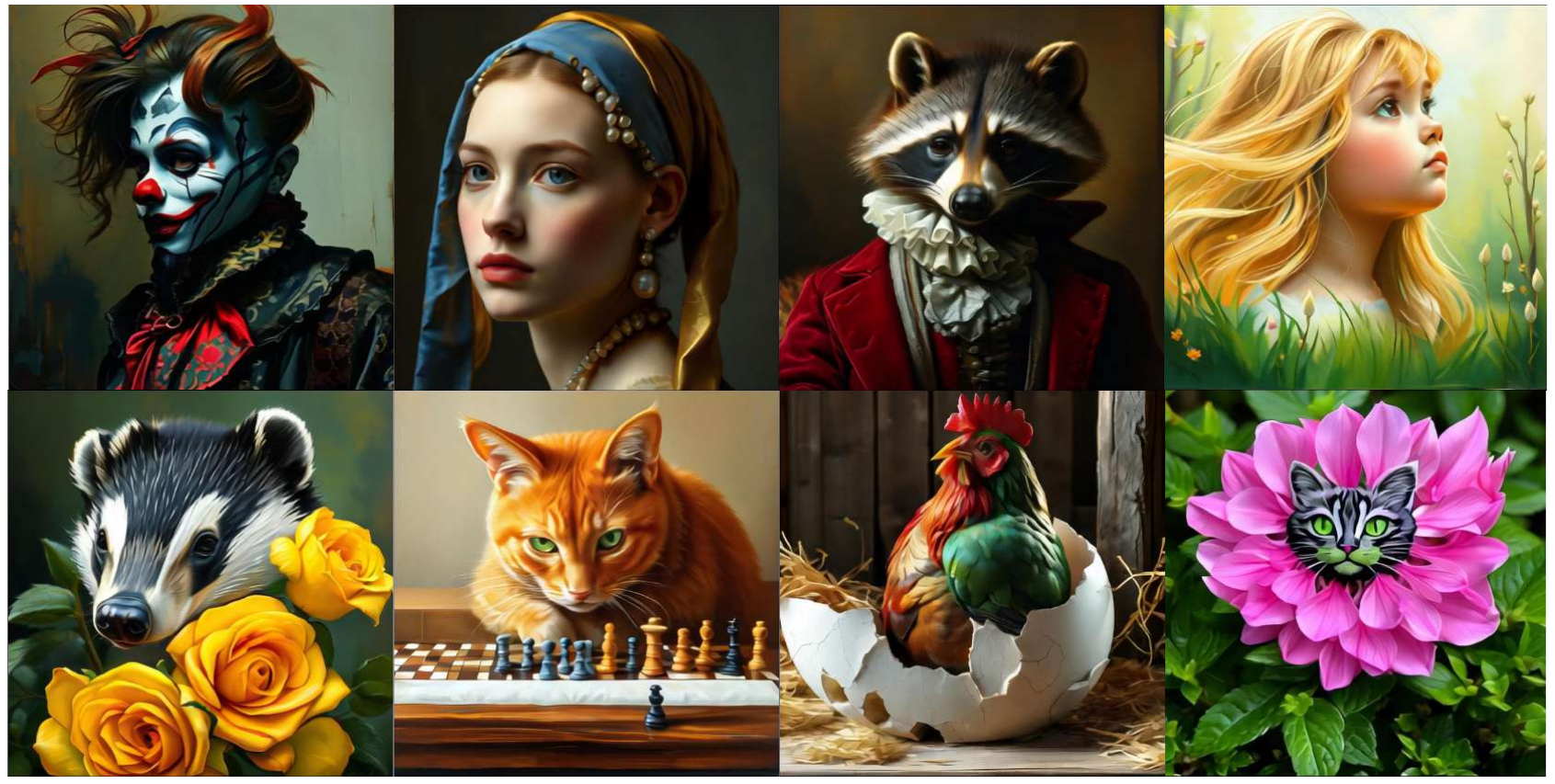}
\vspace{-0.1in}
\caption{Text-to-image generation results by \system in 1024$\times$1024 resolution.}
\label{fig:generation}
\end{figure*}

\begin{abstract}
This work presents \system, a vanilla autoregressive visual generation framework without complex architecure modifications. Through careful exploration of training and inference optimization, we demonstrate that: 1) with only \textbf{0.5B} parameters, our model can generate \textbf{1024$\times$1024 resolution} images with high fidelity, and achieve competitive results on challenging text-to-image benchmarks, \eg, \textbf{0.59 on GenEval and 79.66 on DPG}; 2) both supervised fine-tuning (\textbf{SFT}) and Group Relative Policy Optimization (\textbf{GRPO}) training could lead to significant improvements on generation aesthectics and prompt alignment; and 3) when optimized with inference acceleraton techniques like vLLM, the time for \system to generate an 1024$\times$1024 image could be reduced to around \textbf{14 seconds}. By sharing these findings and open-sourcing the code, we hope to reveal the potential of autoregressive visual generation and encourage more participation in this research field.
\end{abstract}

\section{Introduction}
Recent years have seen the rapid advancements of deep generative models~\cite{ho2020denoising, dhariwal2021diffusion, peebles2023scalable, esser2021taming, tian2024visual, gong2025seedream,wang2025ddt}, which offer innovative approaches for the creation of visual content. Among these models, diffusion models\cite{ho2020denoising, song2020denoising} and autoregressive models\cite{llamagen2024, yu2022scaling} have emerged as the leading paradigms.

Diffusion models~\cite{ho2020denoising, song2020denoising} synthesize visual outputs by iteratively refining random noise through a learned denoising process. This manner has been proven highly effective in producing high-fidelity images~\cite{podell2023sdxl, esser2024scaling} and videos~\cite{blattmann2023stable, kong2024hunyuanvideo, yang2024cogvideox, polyak2025moviegencastmedia,chen2025goku}, thus gaining remarkable popularity in the field of multimodal generation.

Another line of work is autoregressive (AR) models~\cite{llamagen2024,yu2022scaling,kondratyuk2023videopoet}, which formulate visual generation as a sequential process, \ie, each pixel or token is generated based on the preceding ones. This autoregressive process naturally excels in precise and coherent prediction, making it particularly effective for tasks that demand fine-grained control. Moreover, AR visual generation models are naturally compatible with diverse modalities (\eg, language and audio), which facilitates native multimodal understanding and generation~\cite{team2024chameleon,wu2024janus,wang2024emu3,jiao2025unitoken}.

Despite these strengths, autoregressive visual generation still underperforms diffusion models currently. Several hypotheses have been proposed to explain this gap. One posits that discrete visual tokenizer imposes a fundamental limit on the quality of autoregressive generation, while another points to the considerably longer length of visual sequences compared to text, and long-range dependency modeling leads to significantly more challenges. To mitigate these issues, various variants have been proposed, such as MAR~\cite{li2024autoregressive} and VAR~\cite{tian2024visual}. Although these approaches have achieved promising results on academic benchmarks~\cite{deng2009imagenet}, they compromise the intrinsic pattern of ``next-token prediction'' in language models, resulting in a trade-off between performance and simplicity.

In this work, we aim to push the frontier of autoregressive visual generation by preserving the simplicity of the autoregressive framework while carefully optimizing its training paradigm through pretraining, supervised fine-tuning (SFT) and GRPO~\cite{shao2024deepseekmath}-based reinforcement learning (RL). With this, we show that a vanilla AR model with only 0.5B parameters is capable of generating 1024$\times$1024 images with superior aesthetics, and achieving competitive results on existing text-to-image benchmarks, \eg, 0.59 on GenEval. When scaled with more computes (\ie, parameters and tokens), our model consistently demonstrates improved generation quality with higher fidelity and more coherent structures. These results highlight the potential of autoregressive visual generation to compete with diffusion models. 

In addition, we also investigate the inference acceleration of autoregressive visual generation models using vLLM~\cite{kwon2023efficient} and speculative sampling~\cite{chen2023accelerating}. When deployed with vLLM, our model can generate a 1024$\times$1024 image in approximately 14 seconds, showcasing its potential for real-world applications.

\section{Related Work}
\label{sec:related}
\subsection{Autoregressive Visual Generation Models}
Similar to language models~\cite{radford2018improving,touvron2023llama}, autoregressive visual generation methods formulate image and video generation as a next-token prediction process. These approaches rely on
a tokenizer~\cite{esser2021taming,wang2024omnitokenizer,shi2024taming,tian2024reducio,wang2024larp} to tokenize visual inputs into discrete tokens, and train an autoregressive transformer to model the sequential dependencies among these tokens using causal attention. Representative work, including DALL-E~\cite{ramesh2021zero}, Parti~\cite{yu2022scaling}, and LlamaGen~\cite{llamagen2024}, exhibit strong instruction-following and high-fidelity image generation capabilities.

\subsection{Unified Models for Multimodal Understanding and Generation}
Recent advancements in multimodal learning have led to the development of unified models that seamlessly integrate vision and language for both understanding and generation~\cite{wu2024janus,chen2025janus,wu2024vila,wu2024liquid,ma2025unitok}. Chameleon~\cite{team2024chameleon} and Emu3~\cite{wang2024emu3} adopt token-based architecture by handling diverse modalities with a unified autoregressive transformer. Transfusion~\cite{zhou2024transfusion} and Show-o~\cite{xie2024show}, on the other hand, integrate next-token prediction for text and diffusion processes for images within a single model, enabling effective handling of both discrete and continuous data. These unified approaches demonstrate the potential of leveraging large-scale transformer architectures to enhance the synergy between vision and language, setting the stage for more versatile and generalizable multimodal AI systems.

\section{Method}
\label{sec:method}

Our goal is to advance autoregressive visual generation by preserving the simplicity of the AR framework while optimizing its training pipeline and inference efficiency. To this end, we propose \system, which consists of a pretrained visual tokenizer~\cite{agarwal2025cosmos} that discretizes images into compact visual tokens, a text tokenizer, and a decoder-only transformer that autoregressively models the joint distribution of text and image tokens. Note that, unlike diffusion models~\cite{chen2023pixart} or previous AR models~\cite{llamagen2024} that require an additional text encoder~\cite{raffel2020exploring}, our model integrates text encoding and visual generation within a unified transformer architecture, gaining remarkable advantages in both efficiency and multimodal coherence.

\subsection{Autoregressive Visual Generation}

Given an input image $X \in \mathbb{R}^{H \times W \times 3}$, we first tokenize it into a sequence of discrete visual tokens $Z \in \mathbb{R}^{h \times w}$ using a learned tokenizer~\cite{agarwal2025cosmos}, where $p = H/p$ denotes the compression ratio. Each item in $Z$ represents an index of the codebook, mapping a visual patch to a learned discrete representation. Then, these image tokens are flatten into a 1D sequence $z$ in raster scan ordering, following previous work~\cite{esser2021taming,llamagen2024}. We also have a text tokenizer to convert text prompts into text tokens $t \in \mathbb{R}^{N}$.

After this, we concatenate text and image tokens along the sequence dimension, and input them to a transformer decoder~\cite{yang2024qwen2} to model their dependency.

\subsection{Three-stage Training}
\label{subsec:training}
To improve the generation quality and training efficiency, we employ a three-stage learning paradigm: large-scale pretraining on diverse visual datasets to capture generalizable patterns, supervised fine-tuning (SFT) on high-quality data to enhance fidelity and instruction-following, and reinforcement learning (RL) to further refine multimodal alignment~\cite{lee2023aligning,black2023training} and alleviate exposure bias~\cite{ranzato2015sequence,bahdanau2016actor}.

\noindent \textbf{Pretraining and SFT}: during pretraining and supervised-finetuning (SFT), our model is both trained with language modeling loss:
\begin{equation}
    \mathcal{L}_{LM} = - \sum_{i=1}^{L} \mathrm{log} p (z_{i} | z_{<i}, t),
\end{equation}
where the prediction of $z_{i}$ is conditioned on both text tokens $t$ and its preceeding visual tokens $z_{<i}$. 

\noindent \textbf{RL with GRPO}: recently, reinforcement learning has gained increasing attention in the post-training of large language models (LLMs)~\cite{shao2024deepseekmath,huang2025vision} to improve reasoning capability and diffusion models~\cite{lee2023aligning,wallace2024diffusion} for better alignment with human preferences. Among these approaches, Group Relative Policy Optimization (GRPO)~\cite{shao2024deepseekmath} has emerged as a particularly promising technique by offering better training efficiency and stability. 

Using the checkpoint after SFT, GRPO initializes a trainable policy model $\pi_{\theta}$ and a frozen reference model $\pi_{ref}$. For a given text prompt $t$, it samples a group of outputs ${o_{1}, o_{2}, ..., o_{G}}$ from the policy model $\pi_{\theta_{old}}$, and maximizes the following object to optimize $\pi_{\theta}$:

\begin{equation}
    \mathcal{J}_{GRPO}(\theta) = \mathbb{E}_{o_{i} \sim \pi_{\theta_{old}}} \left[
    \begin{aligned}
        &\frac{1}{G}\sum_{i=1}^{G} \mathrm{min} \left( \frac{\pi_{\theta}(o_{i} | t)}{\pi_{\theta_{old}}(o_{i} | t)} A_i, \mathrm{clip} \left( \frac{\pi_{\theta}(o_{i} | t)}{\pi_{\theta_{old}}(o_{i} | t)}, 1 - \epsilon, 1 + \epsilon \right) A_i \right) \\
        &\hspace{5cm} -\beta \mathrm{D}_{KL} (\pi_{\theta} || \pi_{ref})
    \end{aligned}
    \right]
\end{equation}
where $\epsilon$ and $\beta$ denote clipping hyper-parameter~\cite{schulman2017proximal} and coefficient controlling the  Kullback–Leibler (KL) penalty, and $A_{i}$ is the advantage computed in one group. We adopt CLIP~\cite{radford2021learning} as the reward and find it surprisingly useful during experiments.

\subsection{Inference}
During inference, we sample the visual tokens sequentially through the following conditional probability: $\hat{z_{i}} = \mathrm{argmax} p_{\theta} (z_i | z_{<i}, t)$. The sampled tokens are then input to the decoder of visual tokenizer to generate images. Unless otherwise stated, greedy search is employed with $\mathrm{top}K$ = 64000 (codebook size). We also use Classfier-Free Guidance (CFG)~\cite{ho2022classifier} to improve the generation quality.

Since token prediction in autoregressive (AR) models must be performed sequentially, inference latency can be a significant bottleneck. However, various optimizations developed in the LLM community for inference accelerate, such as KV cache~\cite{pope2023efficiently} and paged attention~\cite{kwon2023efficient}, offer promising solutions to this. In this work, we explore the application of these techniques to accelerate AR-based visual generation, which will be introduced below beriefly.

\noindent \textbf{KV Cache}~\cite{pope2023efficiently} stores previously computed key-value embeddings from the attention layers and, reuse them across autoregressive decoding steps to reduce redundant computation. It is widely used in LLM inference and could decrease the complexity from $\mathcal{O}(N^2)$ to $\mathcal{O}(N)$.

\noindent \textbf{vLLM Serving}~\cite{kwon2023efficient} leverages optimized memory management and efficient attention mechanisms, such as paged attention, to enable high-throughput and low-latency inference for autoregressive models on modern hardware.

\noindent \textbf{Speculative Jacobi Decoding}~\cite{chen2023accelerating,teng2024accelerating} accelerates autoregressive generation by first sampling multiple candidate token sequences from a draft model and then efficiently verifying them using the target model. We use the pretrained AR model to serve as both draft model and target model for acceleration.

\section{Experiments}
\subsection{Implementation details}
\noindent \textbf{Experimental Setup}: we adopt the same architecture as Qwen~\cite{bai2023qwen,yang2024qwen2} for our transformer model, and intialize it with LLM weights. While for the visual tokenizer, we use Cosmos-Tokenizer~\cite{agarwal2025cosmos}, whose codebook size is 64k and downsample ratio is 16. 

\begin{table*}[t]
\caption{Evaluation on the GenEval~\cite{ghosh2024geneval} and DPG~\cite{DPG-bench} benchmark. $\dagger$ result is with prompt rewriting. Here we only classify next-token prediction models as AR methods, and next-scale prediction model Infinity as NAR (non-autoregressive) methods.}
\vspace{0.05in}
\small

\centering
\setlength{\tabcolsep}{0.pt}
\renewcommand{\arraystretch}{1.3}
\begin{tabular*}{\linewidth}{@{\extracolsep{\fill}}lccccccccc @{}}
\toprule
\multirow{2}{*}{\textbf{Methods}} & \multirow{2}{*}{\# \textbf{Params}} & \multirow{2}{*}{\textbf{Type}} & \multicolumn{4}{c}{\textbf{GenEval}$\uparrow$} & \multicolumn{3}{c}{\textbf{DPG}$\uparrow$} \\
\cmidrule(l){4-7}\cmidrule(l){8-10}
& & & Two Obj. & Position & Color Attri. & Overall & Global & Relation & Overall \\
\midrule
SDv1.5~\cite{rombach2022high} & 0.9B & {Diff.} & 0.38 & 0.04 & 0.06 & 0.43 &  74.63  & 73.49  &   63.18 \\
PixArt-alpha~\cite{chen2023pixart} & 0.6B & {Diff.} & 0.50 & 0.08 & 0.07 & 0.48 &  74.97  & 82.57  &   71.11 \\
SDv2.1~\cite{rombach2022high} & 0.9B & {Diff.} & 0.51 & 0.07 & 0.17 & 0.50 & 77.67   & 80.72 &   68.09 \\
LlamaGen~\cite{llamagen2024} & 0.8B & {AR} & 0.34 & 0.07 & 0.04 & 0.32 & - & - & 65.16 \\
Ours & 0.5B & {AR} & \textbf{0.82} & \textbf{0.26} & \textbf{0.38} & \textbf{0.59} & \textbf{86.64} & \textbf{88.51} & \textbf{79.66} \\
\midrule
LDM~\cite{rombach2022high} & 1.4B & {Diff.} & 0.29 & 0.02 & 0.05 & 0.37 & - & - & - \\
DALL-E 2~\cite{dalle2} & 6.5B & {Diff.} & 0.66 & 0.10 & 0.19 & 0.52 & - & - & - \\
DALL-E 3~\cite{DALLE3} & - & {Diff.} & - & - & - & 0.67$^{\dagger}$ &  90.97  & 90.58  & \textbf{83.50} \\
Show-o~\cite{xie2024show} & 1.3B & {NAR} & 0.80 &  0.31 &  0.50 & 0.68 & - &- & 67.48 \\
Infinity~\cite{han2024infinity} & 2B & {NAR} &  0.85$^{\dagger}$ & 0.49$^{\dagger}$ & \textbf{0.57}$^{\dagger}$ & \textbf{0.73}$^{\dagger}$ & \textbf{93.11} & \textbf{90.76} & 83.46 \\
Chameleon~\cite{team2024chameleon} & 7B & {AR} & - & - & - & 0.39 & - &- & - \\
Janus~\cite{wu2024janus} & 1.5B & {AR} & 0.68 &  0.46 &  0.42 & 0.61 &  82.33 & 85.46 &  79.68 \\
Emu3~\cite{wang2024emu3} & 8.5B & {AR} & 0.81$^{\dagger}$ & 0.49$^{\dagger}$ & 0.45$^{\dagger}$ & 0.66$^{\dagger}$ & - & - & 81.60 \\
Ours & 1.5B & {AR} & \textbf{0.90} & 0.28 & 0.45 & 0.63 & 87.97 & 88.33 & 81.97 \\
\bottomrule
\end{tabular*}
\label{tab:benchmark}
\vspace{-0.1in}
\end{table*}

As mentioned in Sec.~\ref{subsec:training}, the training consists of three stages: 1) 512 resolution pretraining, 2) 1024 resolution SFT, and 3) 1024 resolution RL. During pretraining / SFT, the learning rate is set to 1e-4 / 2e-5, and batch size is 256 in total. The RL training is conducted using trl~\cite{vonwerra2022trl} framework, the learning rate is 1e-5, and batch size is 28. We do not use warm up and learning rate decay in all stages. AdamW~\cite{loshchilov2017decoupled} is employed for optimization. All experiments are conducted on 32 NVIDIA A100 GPUs.

\noindent \textbf{Training data}: the pretraining data involves CC3M~\cite{sharma2018conceptual}, CC12M~\cite{changpinyo2021conceptual}, OpenImages~\cite{kuznetsova2020open}, SAM1B~\cite{kirillov2023segment}, and Megalith-huggingface~\cite{mega} (around 43M in total). For SFT, we use JourneyDB~\cite{sun2023journeydb}, Synthetic-dataset-1M~\cite{sync}, and 10M internal data. We adopt a simple data filtering strategy for supervised fine-tuning (SFT) data by removing all images whose short edge is smaller than 1024 pixels. We recaption all the images using Qwen2-VL~\cite{wang2024qwen2}, and randomly choose from long and short prompts during training.

\subsection{Comparing with State-of-the-arts}
We present a comprehensive comparison between \system and existing state-of-the-art visual generation models in Table~\ref{tab:benchmark}. Remarkably, with only 0.5B parameters, our model achieves superior performance on established text-to-image benchmarks, \ie, 0.59 overall score on GenEval~\cite{ghosh2024geneval} and 79.66 on DPG-Bench~\cite{DPG-bench}, outperforming all comparable-scale methods (those with fewer than 1B parameters) by significant margins. This includes both diffusion-based approaches (\eg, SDv2.1~\cite{rombach2022high}) and autoregressive alternatives (\eg, LlamaGen~\cite{llamagen2024}). Notably, diffusion models also require auxiliary text encoders (\eg, Flan-T5-XL~\cite{chung2024scaling} with 3B parameters) that effectively double their parameter footprint, but \system process both modalities using a unified transformer, enjoying more efficient parameter utilization and native support for conditional generation.

Moreover, scaling \system to 1.5B yields consistent improvements across benchmarks (+0.04 on GenEval, +1.85 on DPG-Bench), demonstrating predictable scaling behavior analogous to large language models. Although our model still falls behind Infinity~\cite{han2024infinity} on GenEval, we believe this gap is primarily due to the disparity in the volume of training data and can be narrowed with data scaling.

\subsection{Ablation Studies}
In this section, we conduct ablation studies to explore the training and inference of autoregressive visual geneation models, including model initialization, position encodings, \etc.

\begin{table*}[!ht]
\begin{minipage}[t]{0.5\linewidth}
\caption{Initialization of transformer model.}
\centering
\small
\label{tab:ini}
\vspace{0.06in}
\renewcommand{\arraystretch}{1.1}
\setlength{\tabcolsep}{4.0pt}
\begin{tabular*}{0.9\linewidth}{@{\extracolsep{\fill}}lc | ccc@{}}
\toprule
\textbf{Method} && Global & Relation & \textbf{Overall} \\
\midrule
w/o LLM init && 76.10 & 85.61 & 69.43 \\
w/ LLM init && 77.47 & 85.96 & 70.52 \\
\bottomrule
\end{tabular*}
\end{minipage}
\hfill
\begin{minipage}[t]{0.5\linewidth}
\centering
\small
\caption{1D pos embedinggs $v.s.$ 2D.}
\label{tab:pos}
\vspace{0.04in}
\setlength{\tabcolsep}{4.0pt} 
\renewcommand{\arraystretch}{1.1}
\begin{tabular*}{0.9\linewidth}{@{\extracolsep{\fill}}lc | ccc@{}}
\toprule
\textbf{Method} && Global & Relation & \textbf{Overall} \\
\midrule
2D pos && 73.91 & 70.58 & 70.96 \\
1D pos && 77.47 & 85.96 & 70.52 \\
\bottomrule
\end{tabular*}
\end{minipage}
\vspace{-0.05in}
\end{table*}

\noindent \textbf{Effects of LLM initialization}: as shown in Table~\ref{tab:ini}, whether or not using LLM intialization does not have remarkable effect on the DPG-Bench performance. We believe this is due to the text prompts in existing benchmarks is simple and lack sufficient linguistic complexity or reasoning requirements. The results may also indicate that training on text-to-image generation data will lead to drastic forgetting of initial language capabilities. Therefore, incorporating text data during the training phase is essential for maintaining the text understanding and generation capabilities of LLMs.

\noindent \textbf{Effects of Position Encodings}: it is also interesting to comapre 1D and 2D rotary position encoding~\cite{su2024roformer} for image generation using autoregressive framework. We compare the pretraining results on DPG-Bench, and the results in Table~\ref{tab:pos} demonstrate that replacing 1D positional encoding in LLM with 2D will not improve visual generation significantly. However, we believe 2D positional encoding is quite necessary for dynamic resolution generation and video generation.

\noindent \textbf{Effects of RL}: for GRPO-based reinforcement learning, we compare two different reward modules: CLIP-ViT-H-14 and HPSv2~\cite{wu2023human}. Notably, HPSv2 is a fine-tuned version of CLIP-ViT-H-14, trained on a specially constructed human preference dataset. The GenEval performance is quantitatively compared in Table~\ref{tab:rl}, where we observe that using both reward modules results in performance improvements. Specifically, the CLIP reward achieves a greater performance gain, with a +0.6 increase on GenEval for the 0.5B model. The qualitative results, as shown in Figure~\ref{fig:grpo}, further demonstrate that CLIP could enhance the text rendering capability, along with improved perception of quantifiers and spatial descriptions.

We also plot the reward value and GenEval performance during training in Figure~\ref{fig:reward}-~\ref{fig:geneval}. As can be seen from the left, the reward values exhibit a gradual increase during training. Interestingly, the GenEval performance demonstrates a positive correlation with the reward progression, suggesting that a simple reward function, \ie, CLIP-ViT-H-14, is effective in providing consistent feedback that aligns well with the desired task performance.

\begin{table*}[t]
\begin{minipage}[t]{0.45\linewidth}
\centering
\small
\caption{GenEval performance of SFT/RL model, TO: two objects, P: position, CA: color attribute.}
\vspace{0.07in}
\renewcommand{\arraystretch}{1.4}
\setlength{\tabcolsep}{0pt}
\begin{tabular*}{\linewidth}{@{\extracolsep{\fill}}lc|cc|cccc@{}}
\toprule
\textbf{Param} && \textbf{Reward} && \textbf{TO} & \textbf{P} & \textbf{CA} & \textbf{Overall} \\
\midrule
\multirow{3}*{0.5B} && - && 0.73 & 0.22 & 0.23 & 0.53 \\
~ && HPSv2 && 0.81 & 0.23 & 0.32 & 0.56 \\
~ && CLIP && 0.82 & 0.26 & 0.38 & 0.59 \textcolor{red}{($\uparrow$ 0.6)} \\
\midrule
\multirow{3}*{1.5B} && - && 0.87 & 0.27 & 0.33 & 0.61 \\
~ && HPSv2 && 0.94 & 0.26 & 0.38 & 0.61 \\
~ && CLIP && 0.90 & 0.28 & 0.45 & 0.63 \textcolor{red}{($\uparrow$ 0.2)} \\
\bottomrule
\end{tabular*}
\label{tab:rl}
\end{minipage}
\hfill
\begin{minipage}[t]{0.5\linewidth}
\centering
\figcaption{Generation results before/after GRPO.}
\vspace{0.05in}
\includegraphics[width=\linewidth]{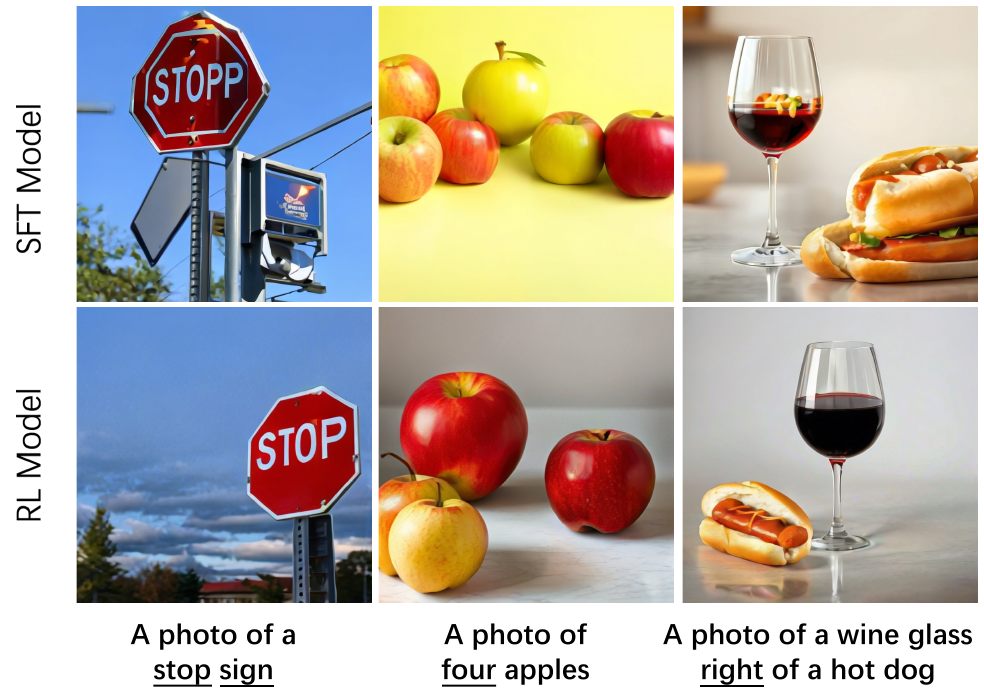}
\label{fig:grpo}
\end{minipage}
\vspace{-0.15in}
\end{table*}

\begin{table*}[!ht]
\begin{minipage}[t]{0.5\linewidth}
\centering
\includegraphics[width=0.95\linewidth]{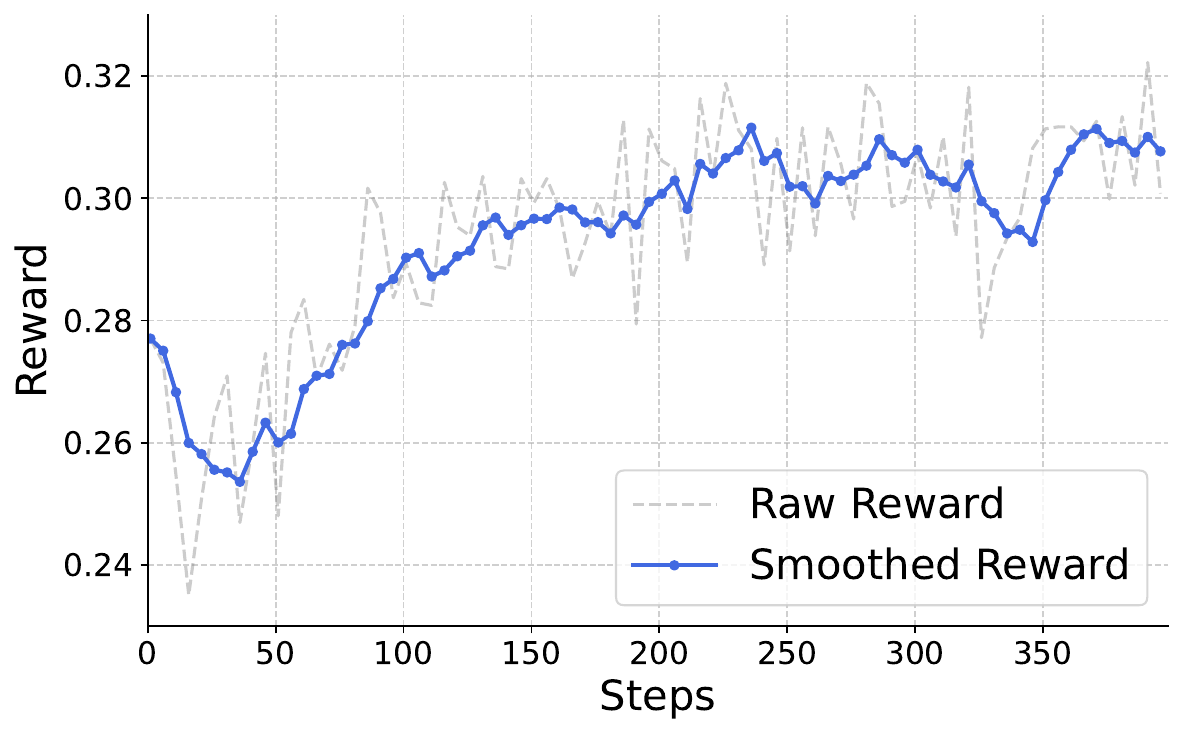}
\label{fig:reward}
\end{minipage}
\hfill
\begin{minipage}[t]{0.5\linewidth}
\centering
\includegraphics[width=0.95\linewidth]{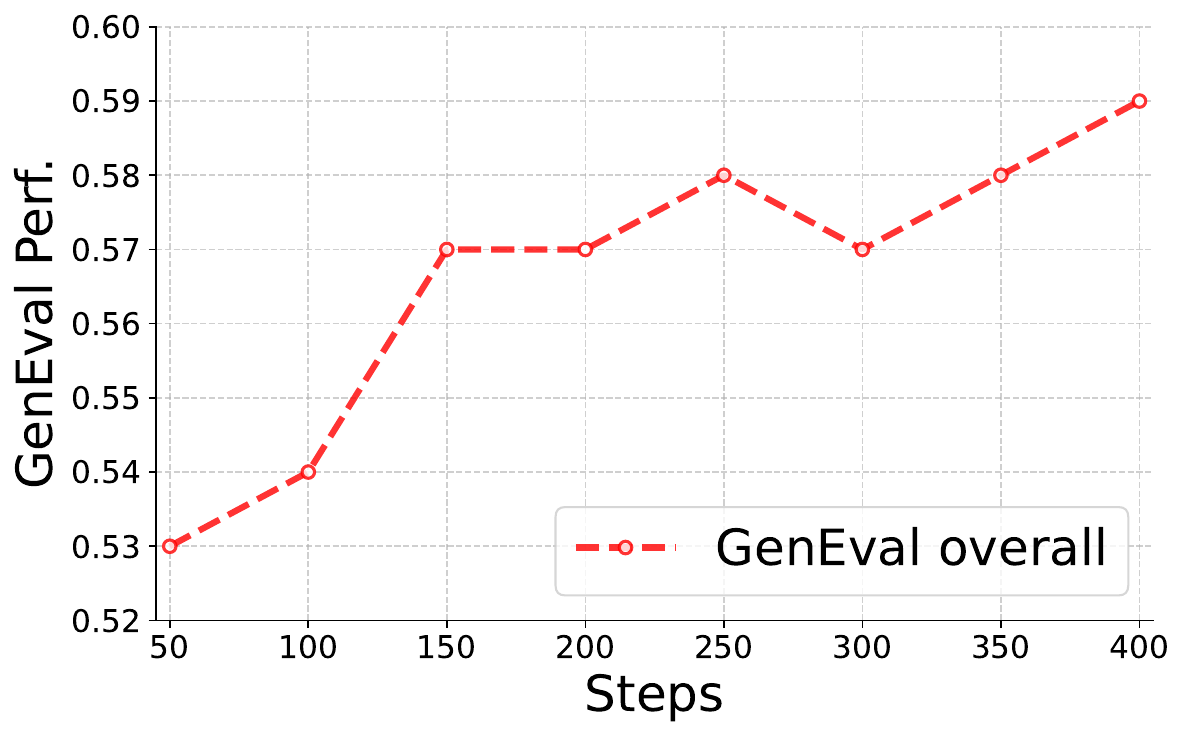}
\label{fig:geneval}
\end{minipage}
\vspace{-0.25in}
\figcaption{Visualizations of the GRPO training: reward and GenEval overall performance.}
\vspace{-0.2in}
\end{table*}

\noindent \textbf{Inference Speedup}: the sequential prediction nature of autoregressive (AR) models could result in high inference latency, making it challenging to deploy them in real-time applications. However, recent advancements in LLM optimization techniques, such as KV cache, paged attention, and speculative decoding, offer opportunities for accelerating inference in AR visual generation models. Therefore, we conduct experiments to apply these approaches to speed up the inference of \system. The throughput is calculated on an Nvidia A100 node, with CFG enabled.

\begin{table*}[!ht]
\begin{minipage}[t]{0.45\linewidth}
\centering
\small
\caption{Inference speed comparison w/ KV Cache and vLLM.}
\label{tab:speed}
\vspace{0.07in}
\renewcommand{\arraystretch}{1.2}
\setlength{\tabcolsep}{4.0pt}
\begin{tabular*}{\linewidth}{lc | c}
\toprule
\textbf{Method} && \textbf{Throughput (sec/img)} \\
\midrule
baseline && 227.62 \\
+ KV Cache && 150.19 \\
+ vLLM && 13.55 \\
\bottomrule
\end{tabular*}
\end{minipage}
\hfill
\begin{minipage}[t]{0.55\linewidth}
\centering
\small
\caption{Speculative Jacobi Decoding.}
\label{tab:sjd}
\vspace{0.04in}
\setlength{\tabcolsep}{0pt} 
\renewcommand{\arraystretch}{1.2}
\begin{tabular*}{0.8\linewidth}{@{\extracolsep{\fill}}lc | cc | c@{}}
\toprule
\textbf{Method} && \textbf{Avg Steps} && \textbf{DPG-overall} \\
\midrule
baseline && 4096 && 79.66 \\
SJD && 2685 && 80.33 \\
SJD-w16 && 2199 && 81.39 \\
SJD-w32 && 2034 && 81.05 \\
\bottomrule
\end{tabular*}
\end{minipage}
\end{table*}

Table~\ref{tab:speed} demonstrates that using KV cache can effectivelly save 34\% inference time, while serving with vLLM can lead to more sigficant inference acceleration, reducing the time to generate a 1024$\times$1024 image to 13.55 sceconds. 

We also try speculative jacobi decoding (SJD), which speculatively decoding multiple tokens in parallel~\cite{song2021accelerating,chen2023accelerating} at inference time to reduce the autoregressive generation steps. The results are shown in Table~\ref{tab:sjd}, we can see that SJD can lead to around 2$\times$ reduction in steps and slightly better performance on DPG. We also compare the sliding-window design proposed by~\cite{teng2024accelerating}. Although SJD does not practically reduce the testing latency of the autoregressive (AR) model (unable to use KV cache and need to forward the entire sequence each time), it still presents many possibilities for optimizing the AR inference process.

\begin{figure}[htbp]
\centering
\includegraphics[width=0.93\linewidth]{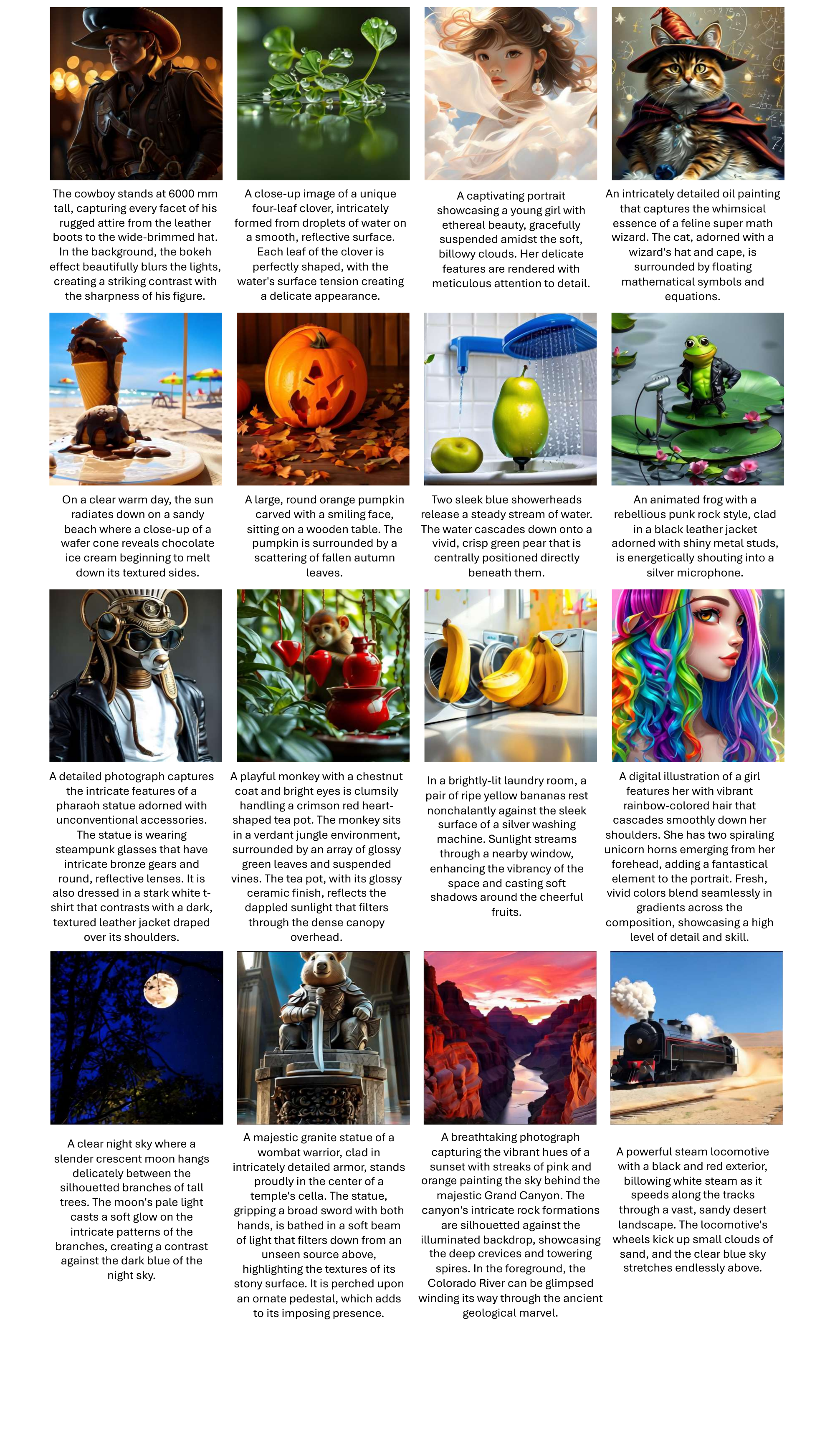}
\vspace{-0.12in}
\caption{Text-to-image generation results by \system using DPG prompts.}
\label{fig:vis}
\end{figure}

\subsection{Visualizations and Failure Cases}
\begin{figure*}[!t]
\centering
\includegraphics[width=0.98\linewidth]{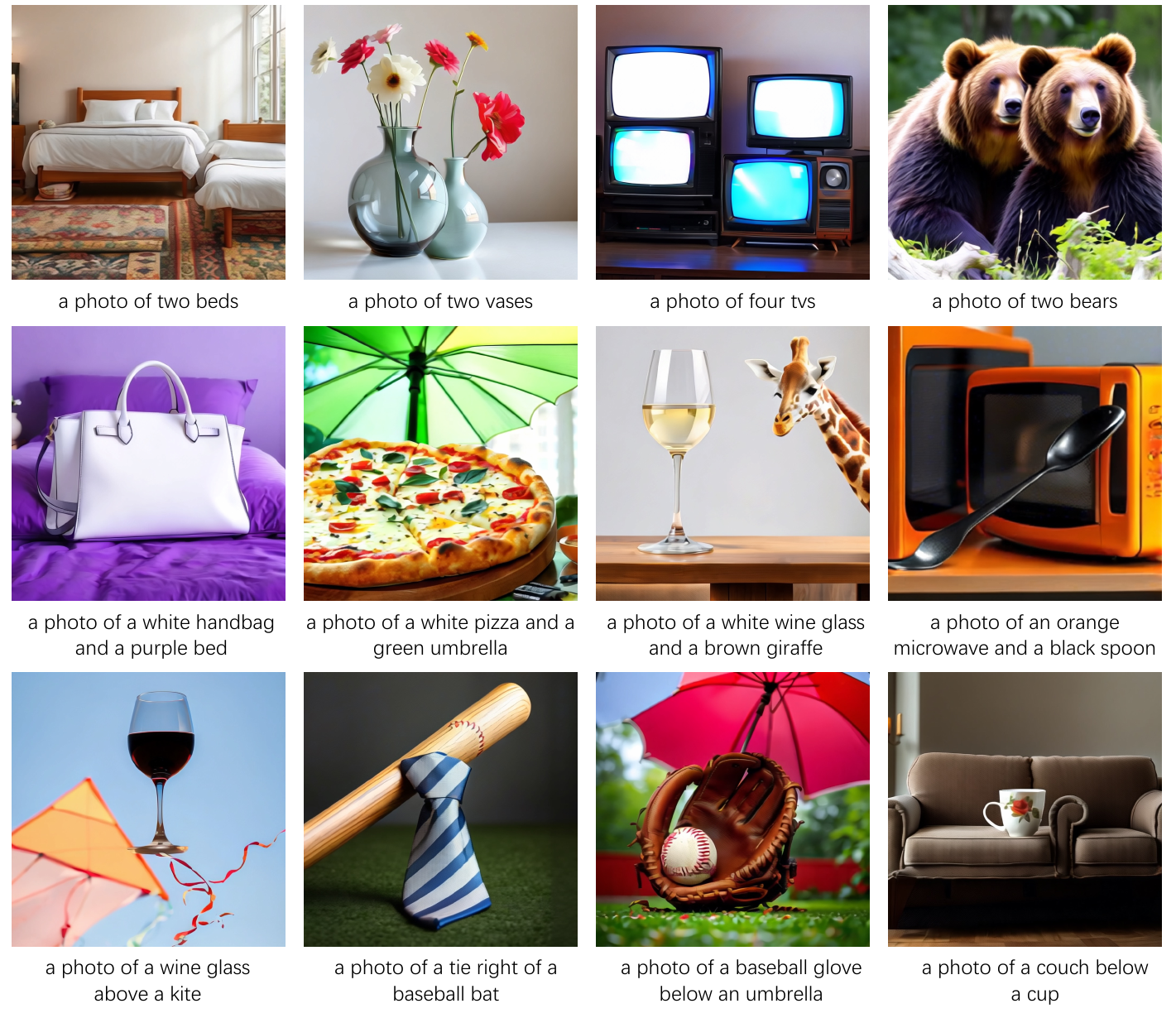}
\vspace{-0.1in}
\caption{Text-to-image generation results by \system using GenEval prompts.}
\label{fig:vis2}
\vspace{-0.05in}
\end{figure*}

We visualize image generation results of \system in Figure~\ref{fig:vis} and~\ref{fig:vis2}. It can be observed that our model could not only generate high-fidelity, aesthetically pleasing images but also demonstrate strong instruction-following capabilities.

\begin{figure*}[!ht]
\centering
\vspace{-0.15in}
\includegraphics[width=0.98\linewidth]{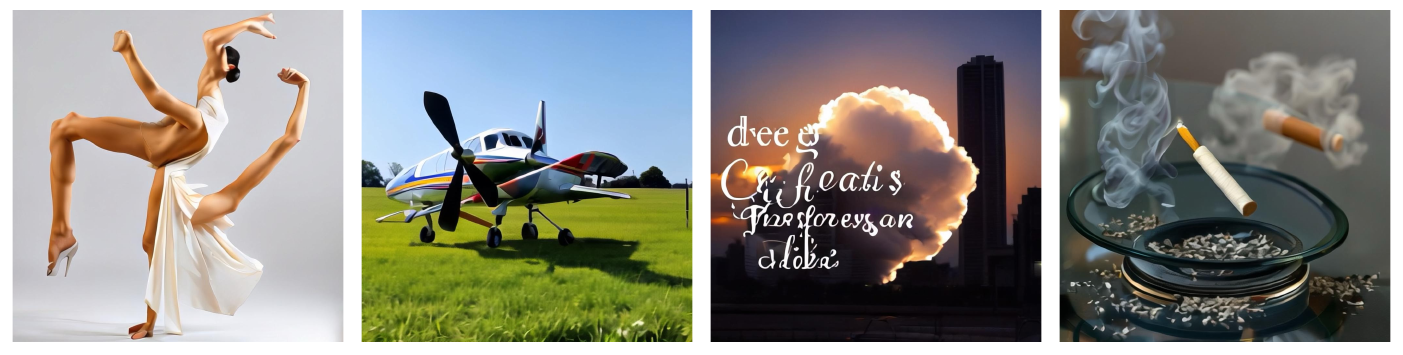}
\vspace{-0.1in}
\caption{Failure cases of \system.}
\label{fig:failure}
\vspace{-0.1in}
\end{figure*}

Several failure cases are also shown in Figure~\ref{fig:failure}. The limited data scale and parameter size constrain \system to generate complex poses, objects, and text. Additionally, our model may synthesize content that does not adhere to physical laws.

\section{Conclusion and Future Work}

This work presented \system, a vanilla autoregressive framework for visual generation that discards complex architecture modifications. We focus on the optimization of two fundamental components: 1) Training pipeline, through large-scale pretraining, high-quality supervised finetuning, and GRPO training, \system achieves competitive performance on existing text-to-image generation benchmarks with only 0.5B parameters. 2) Inference efficiency, we explore various inference acceleration techniques, and show \system could generate a 1024$\times$1024 image in around 14 seconds when served with vLLM. We hope this work can inspire further exploration into autoregressive visual generation and firmly believe that it is a promising alternative to diffusion models. 

Despite the superior results achieved, we admit that there still exist many limitations that are worth deeper improvements or exploring:

\noindent \textbf{Stronger visual tokenizers}: the reconstruction performance of Cosmos-Tokenizer~\cite{agarwal2025cosmos} is limited, especially in capturing fine-grained visual details, \eg, faces and texts. This leaves room for better visual generation results with improved tokenization methods.

\noindent \textbf{Text-to-video generation}: compared to text-to-image generation, text-to-video generation presents significantly more challenges since the model has to generate coherent outputs that are contextually and temporally consistent.

\noindent \textbf{Native multimodal understanding and generation}: recently, the native multimodal understanding and generation capabilities of GPT-4o have captured widespread attention. It offers a glimpse into the future of models that can seamlessly integrate vision, text, and even audio without relying on modality-specific encodings. Moving \system forward, building truly native large multimodal models that can perform end-to-end reasoning across images, text, and other modalities is a promising and crucial research direction.

\bibliographystyle{abbrv}
\bibliography{main}

\end{document}